\documentclass[final,5p,times,twocolumn]{elsarticle}
\usepackage[T1]{fontenc}
\usepackage{amssymb}
\usepackage{amsmath}
\usepackage{graphicx}
\usepackage{float}
\usepackage{booktabs}
\usepackage{threeparttable}
\usepackage{siunitx}
\usepackage{bm}
\usepackage{multirow}
\usepackage{placeins}
\usepackage{booktabs}
\usepackage{multirow}
\usepackage{makecell}
\usepackage{url}
\usepackage{stfloats}
\journal{}

\begin{document}
\frenchspacing

\begin{frontmatter}

\title{AnF-DiffPET: Anatomy- and Frequency-Guided Diffusion for PET/CT Denoising}

\author[inst1]{Xuepeng Liu}
\author[inst2]{Ruili Li}
\author[inst1]{Zetong Liu}
\author[inst1]{Renyiming Li}
\author[inst1]{Yan Li}
\author[inst1]{Yin Dai}
\author[inst3]{Chao Li}
\author[inst1]{Yueyang Teng\corref{cor1}}
\ead{tengyy@bmie.neu.edu.cn}
\cortext[cor1]{Corresponding author}

\affiliation[inst1]{organization={College of Medicine and Biological Information Engineering, Northeastern University},
                   city={Shenyang 110169},
                   country={China}}
\affiliation[inst2]{organization={Tohoku University School of Medicine},
                   city={Sendai 980-8575},
                   country={Japan}}
\affiliation[inst3]{organization={Department of Clinical Neurosciences, University of Cambridge},
                   country={UK}}

\begin{abstract}
Positron emission tomography (PET) provides essential functional information for disease assessment, however reducing injected activity or acquisition time produces low-dose (LD) PET with stronger count dependent noise and less reliable uptake quantification. Diffusion models offer a promising solution for PET denoising by progressively recovering high-dose (HD) PET images from LD inputs. However, LD-to-HD PET denoising is still challenging due to insufficient anatomical guidance, unstable multi-scale feature propagation, and uncertain frequency domain uptake recovery. We propose AnF-DiffPET, an anatomy- and frequency-guided diffusion framework for computed tomography (CT) conditioned LD PET denoising. The framework integrates Anatomical-Frequency Guidance (AFG), Multi-Scale Cross-Transformer Reconstruction (MSCTR), and Frequency-Contrastive Hard Mining (FCHM) to enhance anatomy aware feature modulation and frequency domain consistency during denoising. Experimental results across four PET/CT datasets show that the proposed method improves image fidelity, anatomical consistency, and quantitative fidelity over representative CNN-based, GAN-based, transformer-based, and diffusion-based methods. The code and trained models will be publicly released upon acceptance.

\end{abstract}


\end{frontmatter}

\section{Introduction}
Positron emission tomography (PET) is widely used for oncological diagnosis, treatment response assessment, and quantitative metabolic evaluation.
When combined with computed tomography (CT), PET/CT further provides spatially aligned functional anatomical information, enabling lesion localization and quantitative interpretation in routine clinical imaging~\cite{PET/CT,basu2011fundamentals,pet/ct2}.
However, PET image quality is strongly affected by the number of detected photon counts.
Reducing injected radiotracer activity or shortening acquisition time can reduce radiation burden and improve clinical efficiency, however, it also produces low-dose (LD) PET with stronger count dependent noise, reduced signal-to-noise ratio, and less stable uptake estimates.
These degradations make it difficult to delineate weak or small uptake structures and can reduce the reliability of PET quantification~\cite{tian2024deep,tivnan2025generative}.
Recent PET denoising work similarly defines this task as recovering high-dose (HD) PET images from LD counterparts acquired with reduced tracer dose or shorter scanning duration, where limited photon counts increase noise and reduce signal-to-noise ratio~\cite{yang2026unipet}.
Denoising LD PET to recover HD PET appearance while preserving uptake fidelity is therefore a central challenge in LD PET/CT denoising.

Conventional methods address PET degradation either during image formation or through post processing, for example by incorporating statistical models into reconstruction, applying deconvolution, or using denoising and detail enhancing filters~\cite{Pain_2022_03,alessio2006modeling,cloquet2010non}.
These methods can suppress noise or sharpen images, however, they often depend on simplified degradation assumptions.
As a result, they may over-smooth focal uptake, amplify residual noise, or introduce artifacts, making it difficult to improve image appearance while preserving quantitatively stable uptake estimates.

Deep learning has therefore been widely explored for PET image denoising.
Existing methods commonly formulate LD PET denoising as a direct LD-to-HD image recovery problem, using convolutional networks such as VDSR~\cite{vdsr} and EDSR~\cite{edsr} or adversarial models such as ESRGAN~\cite{esrgan} to map degraded PET directly to HD PET outputs.
Although these methods can improve image appearance and suppress visible noise, their deterministic regression objectives provide limited modeling of degradation uncertainty and spatially varying PET noise.
This limitation can lead to over-smoothed uptake boundaries or visually plausible details that are unreliable for quantitative interpretation~\cite{chen2022spatial,cui2024pet}.
Transformer-based denoising methods, represented by SwinIR~\cite{swinir}, use shifted window self-attention to enlarge the effective receptive field and improve long range feature interaction, however, their deterministic reconstruction objective may still smooth weak uptake or hallucinate texture when the LD PET input contains strong count dependent noise.

Diffusion models provide an alternative by formulating image recovery as progressive denoising~\cite{ddpm,pan20232d}.
In natural and medical imaging, diffusion-based denoising methods such as I2SB~\cite{I2SB}, and RDDM~\cite{rddm} often recover finer details and more stable structures than direct regression models.
Diffusion-based LD PET/CT denoising, however, remains underexplored in three connected respects.
First, existing methods often apply diffusion directly to the LD PET image, which limits their ability to preserve anatomical consistency and recover fine structural details during reconstruction. Second, conditional guidance is often static or coarse, allowing LD noise propagation and structural drift to accumulate across diffusion steps. Third, under count limited degradation, the frequency domain distribution of PET uptake becomes uncertain and requires explicit spectral regulation.
Together, these factors create a failure chain in which insufficient anatomical conditioning leaves the structural target ambiguous, coarse guidance allows stepwise drift, and weak frequency control destabilizes HD PET uptake recovery.

To address these limitations, we propose AnF-DiffPET, an anatomy- and frequency-guided diffusion framework for CT conditioned LD PET denoising.
Rather than applying diffusion directly to degraded PET images with limited conditioning, AnF-DiffPET integrates coregistered CT priors into the denoising process and jointly enhances structural prior modeling, multi-scale feature consistency, and frequency domain regularization.
Specifically, the Anatomical-Frequency Guidance (AFG) module adaptively injects CT derived anatomical cues into the encoder to steer PET feature extraction toward anatomically consistent representations. The Multi-Scale Cross-Transformer Reconstruction (MSCTR) module replaces direct decoder skip fusion with attention based multi-scale feature reconstruction to reduce noisy feature propagation and structural drift. The Frequency-Contrastive Hard Mining (FCHM) strategy constrains spectral recovery during training to suppress noise dominated frequency responses under LD-to-HD denoising. We evaluate AnF-DiffPET on HECKTOR 2025 and three additional PET/CT datasets spanning prostate molecular imaging, bone focused uptake, and thoracic lung cancer imaging. Extensive comparisons with representative CNN-based, GAN-based, transformer-based, and diffusion-based methods demonstrate that AnF-DiffPET improves image fidelity, anatomical consistency, and quantitative fidelity across diverse imaging scenarios.

\section{Methodology}

\subsection{Overview}

\begin{figure*}[!tbp]
    \centering
    \includegraphics[width=\textwidth]{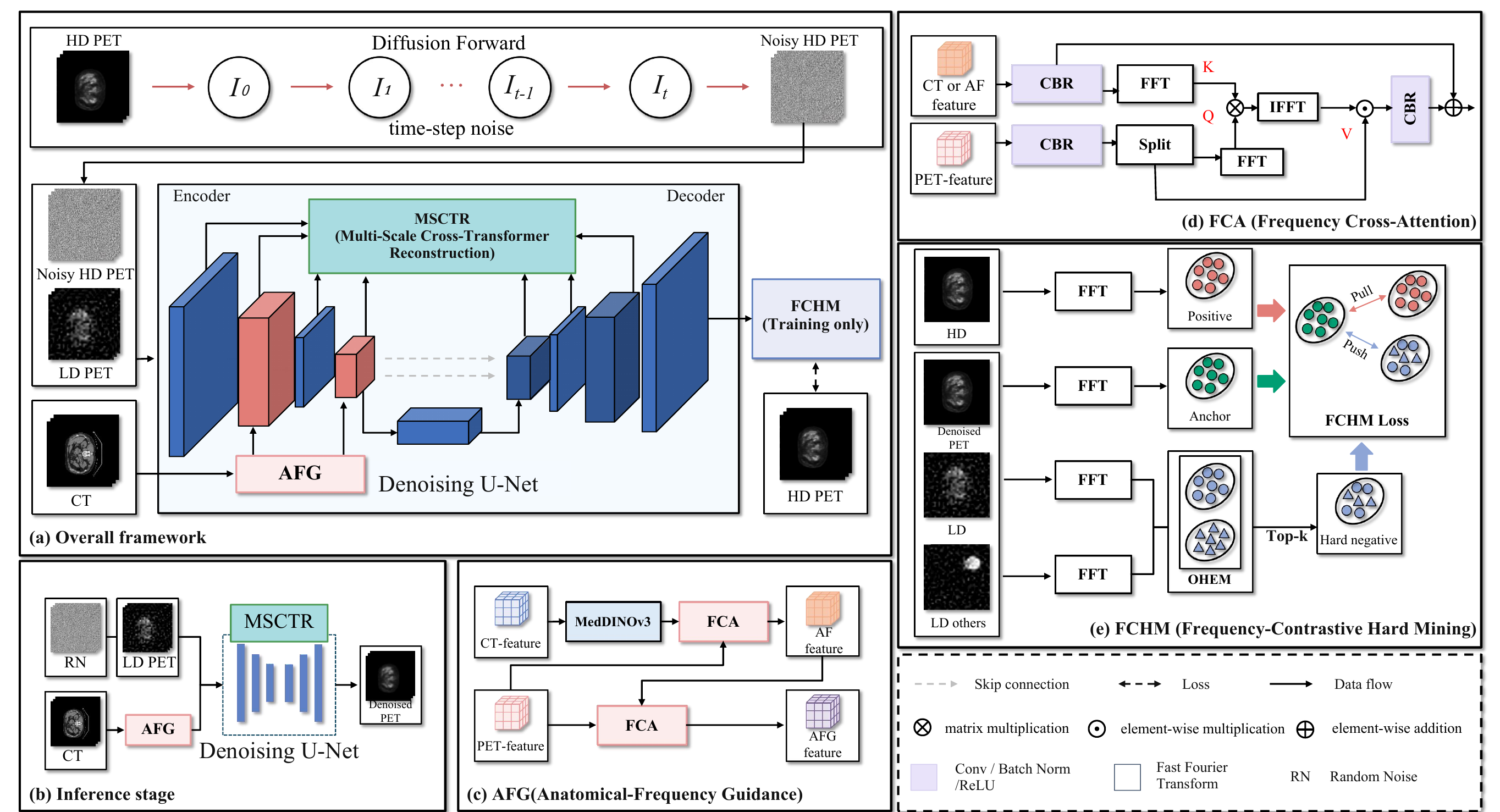}
\caption{Overall framework of AnF-DiffPET for CT conditioned LD PET denoising.
(a) Training workflow with HD PET diffusion and LD PET/CT conditioned denoising.
(b) Inference workflow using only LD PET and CT to generate the denoised HD PET image.
(c) AFG module with cascaded FCA blocks for CT derived anatomical-frequency guidance.
(d) FCA block for frequency domain query-key interaction and spatial domain value modulation.
(e) FCHM for Top-$k$ hard LD negative mining and frequency-contrastive regularization during training.}
    \label{fig:model}
\end{figure*}

To address anatomical ambiguity, low-count feature drift, and spectral inconsistency in LD PET denoising, we formulate AnF-DiffPET as a CT conditioned diffusion denoising framework. 
Fig.~\ref{fig:model} summarizes both the training and inference workflows. 
During training, the model predicts a denoised HD PET image from a noisy HD PET sample under LD PET/CT conditioning, while Frequency-Contrastive Hard Mining (FCHM) is applied as a training-only frequency domain regularizer. 
During inference, the model starts from random noise and uses only LD PET and CT to generate the final denoised HD PET image.

Let $\mathbf{x}_\mathrm{HD}\in\mathbb{R}^{H\times W}$ be the raw image, $\mathbf{y}_{\mathrm{LD}}\in\mathbb{R}^{H\times W}$ be the corresponding LD PET image on the same spatial grid, and $\mathbf{z}\in\mathbb{R}^{H\times W}$ be the CT image spatially registered to PET. 
In the forward process, Gaussian noise is added to $\mathbf{x}_{HD}$ over $T$ diffusion steps:
\begin{equation}
q(\mathbf{x}_t \mid \mathrm{x}_\mathrm{HD})
=
\mathcal{N}
\left(
\sqrt{\bar{\alpha}_t}\,\mathbf{x}_\mathrm{HD},\,
(1-\bar{\alpha}_t)\mathbf{I}
\right)
\end{equation}
where $t\in\{1,\dots,T\}$ is the diffusion time step, $\beta_t$ is the noise variance at step $t$, $\alpha_t=1-\beta_t$, and $\bar{\alpha}_t=\prod_{i=1}^{t}\alpha_i$. 
Equivalently, the noisy HD PET image can be written as
\begin{equation}
\mathbf{x}_t
=
\sqrt{\bar{\alpha}_t}\,\mathbf{x}_\mathrm{HD}
+
\sqrt{1-\bar{\alpha}_t}\,\boldsymbol{\epsilon},
\qquad
\boldsymbol{\epsilon}\sim\mathcal{N}(\mathbf{0},\mathbf{I})
\end{equation}

The reverse process learns $p_{\theta}$ to denoise progressively conditioned on $(\mathbf{y}_{\mathrm{LD}},\mathbf{z})$:
\begin{equation}
p_{\theta}
(
\mathbf{x}_{t-1}
\mid
\mathbf{x}_t,\mathbf{y}_{\mathrm{LD}},\mathbf{z}
)
=
\mathcal{N}
\left(
\mu_{\theta}(\mathbf{x}_t,\mathbf{y}_{\mathrm{LD}},\mathbf{z},t),\,
\Sigma_{\theta}(\mathbf{x}_t,\mathbf{y}_{\mathrm{LD}},\mathbf{z},t)
\right)
\end{equation}
where $\theta$ denotes the learnable parameters. 
In our implementation, the denoising U-Net adopts an $\mathbf{x}_\mathrm{HD}$-prediction parameterization. 
Given the noisy HD PET image $\mathbf{x}_t$, the LD PET condition $\mathbf{y}_{\mathrm{LD}}$, the registered CT image $\mathbf{z}$, and the diffusion step $t$, the network predicts the clean HD PET estimate:
\begin{equation}
\hat{\mathbf{x}}_{\mathrm{HD}}
=
f_{\theta}
\left(
\mathbf{x}_t,\mathbf{y}_{\mathrm{LD}},\mathbf{z},t
\right)
\end{equation}
where $f_{\theta}(\cdot)$ denotes the denoising U-Net. 
The predicted $\hat{\mathbf{x}}_{\mathrm{HD}}$ is used to parameterize the reverse mean as
\begin{equation}
\mu_{\theta}
\left(
\mathbf{x}_t,\mathbf{y}_{\mathrm{LD}},\mathbf{z},t
\right)
=
\frac{
\sqrt{\bar{\alpha}_{t-1}}\beta_t
}{
1-\bar{\alpha}_t
}
\hat{\mathbf{x}}_{\mathrm{HD}}
+
\frac{
\sqrt{\alpha_t}\left(1-\bar{\alpha}_{t-1}\right)
}{
1-\bar{\alpha}_t
}
\mathbf{x}_t
\end{equation}
with $\bar{\alpha}_0=1$. 
The covariance is fixed by the noise schedule as $\Sigma_{\theta}=\tilde{\beta}_t\mathbf{I}$, where
\begin{equation}
\tilde{\beta}_t
=
\frac{1-\bar{\alpha}_{t-1}}{1-\bar{\alpha}_t}\beta_t 
\end{equation}
Thus, the network directly predicts the denoised HD PET image rather than an LD-to-HD residual.

\subsection{Anatomical-Frequency Guidance Module(AFG)}

After defining the conditional diffusion backbone, we describe how anatomical information is introduced into the denoising process. 
Since LD PET contains count dependent noise and weak apparent structural boundaries, the paired CT image provides complementary anatomical cues for PET feature extraction. 
Given the CT image $\mathbf{z}$, we extract multi-scale dense anatomical priors using a frozen pretrained MedDINOv3~\cite{li2025meddinov3} encoder:
\begin{equation}
\mathbf{a}^{\,l}
=
\mathrm{MedDINOv3}^{\,l}(\mathbf{z}),
\qquad
\mathbf{f}^{\,l}_{t}
=
\text{U-Net}^{\,l}
\left(
\mathbf{y}_{\mathrm{LD}},\,\mathbf{x}_{t},\,t
\right)
\end{equation}
where $\mathbf{a}^{\,l}$ denotes the anatomical feature at scale $l$, aligned with encoder stage $l$ of the denoising U-Net, and $\mathbf{f}^{\,l}_{t}$ denotes the PET feature extracted from the LD PET condition $\mathbf{y}_{\mathrm{LD}}$ and the noisy HD PET image $\mathbf{x}_{t}$.

The extracted PET feature $\mathbf{f}^{\,l}_{t}$ is fused with the anatomical prior $\mathbf{a}^{\,l}$ through the proposed AFG using cascaded Frequency Cross-Attention (FCA) blocks. 

The first FCA block produces an anatomical-frequency feature (AF feature), and the second FCA block uses this AF feature to produce anatomical-frequency feature guidance feature (AFG feature) to further guide the original PET representation:
\begin{equation}
\mathbf{u}^{\,l}_{t}
=
\mathrm{FCA}
\left(
\mathbf{a}^{\,l},\,\mathbf{f}^{\,l}_{t}
\right),
\qquad
\mathbf{f}^{\,l,\mathrm{out}}_{t}
=
\mathrm{FCA}
\left(
\mathbf{u}^{\,l}_{t},\,\mathbf{f}^{\,l}_{t}
\right) 
\end{equation}
Here, $\mathbf{u}^{\,l}_{t}$ corresponds to the AF feature shown in Fig.~\ref{fig:model}(c), and $\mathbf{f}^{\,l,\mathrm{out}}_{t}$ denotes the AFG feature forwarded to subsequent U-Net stages.

Within each FCA block, the AF feature is injected into PET features through frequency domain query-key interaction and spatial domain value modulation. 
For stage $l$ of the denoising U-Net, let the AF feature  and PET feature be denoted as $\mathbf{g}^{\,l}\in\mathbb{R}^{H_l\times W_l\times C_g}$ and $\mathbf{f}^{\,l}_{t}\in\mathbb{R}^{H_l\times W_l\times C_f}$, respectively. 
In the first FCA block, $\mathbf{g}^{\,l}=\mathbf{a}^{\,l}$; in the second FCA block, $\mathbf{g}^{\,l}=\mathbf{u}^{\,l}_{t}$. 
We first apply a Conv-BN-ReLU (CBR) block for channel alignment and local refinement. 
Here, CBR denotes a $3 \times 3$ convolutional layer followed by batch normalization and ReLU activation. Then, we split the refined PET feature along the channel dimension:
\begin{equation}
\bar{\mathbf{f}}^{\,l}_{t}
=
\mathrm{CBR}
\left(
\mathbf{f}^{\,l}_{t}
\right),
\qquad
\bar{\mathbf{f}}^{\,l}_{t}
=\mathrm{Split}
\left[
\bar{\mathbf{f}}^{\,l,(q)}_{t},
\bar{\mathbf{f}}^{\,l,(v)}_{t}
\right] 
\end{equation}
where $\bar{\mathbf{f}}^{\,l,(q)}_{t}$ is used to construct the query branch and $\bar{\mathbf{f}}^{\,l,(v)}_{t}$ is preserved as the spatial domain value branch. Split denotes the channel-wise partitioning of the feature into two parts, serving as the query and value branches, respectively.

We construct the query, key, and value representations using separate parameterized projections, where the query and key are computed in the Fourier domain via the fast Fourier transform (FFT), while keeping the value in the spatial domain:
\begin{equation}
\begin{aligned}
\mathbf{Q}^{\,l}_{t}
&=
\mathrm{FFT}
\left[
\mathrm{MLP}
\left(
\mathrm{LN}\left(\bar{\mathbf{f}}^{\,l,(q)}_{t}\right)
\right)
\right]),\\
\mathbf{K}^{\,l}
&=
\mathrm{FFT}
\left[
\mathrm{MLP}
\left(
\mathrm{LN}\left(\mathbf{g}^{\,l}\right)
\right)
\right],\\
\mathbf{V}^{\,l}_{t}
&=
\mathrm{MLP}
\left[
\mathrm{LN}\left(\bar{\mathbf{f}}^{\,l,(v)}_{t}\right)
\right]
\end{aligned}
\end{equation}
where MLP denotes a multilayer perceptron, and LN denotes layer normalization.

Since FFT produces complex-valued representations, the attention affinity is computed on the magnitude spectra of the query and key. 
The frequency domain affinity is computed and mapped back to the spatial domain via the inverse fast Fourier transform (IFFT):
\begin{equation}
\mathbf{A}^{\,l}_{t}
=
\mathrm{Softmax}
\left(
\frac{
\left|
\mathbf{Q}^{\,l}_{t}
\right|
\left|
\mathbf{K}^{\,l}
\right|^{\top}
}{
\sqrt{d}
}
\right),
\qquad
\mathbf{M}^{\,l}_{t}
=
\mathrm{Re}
\left[
\mathrm{IFFT}
\left(
\mathbf{A}^{\,l}_{t}
\right)
\right]
\end{equation}
where $d$ is the feature dimension for normalization. $\otimes$ denotes the Matrix multiplication, and $\mathrm{Re}[\cdot]$ takes the real component after IFFT.

The resulting spatial attention map $\mathbf{M}^{\,l}_{t}$ modulates the value branch and is refined by a CBR block. 
The output is then fused with the PET feature through a residual connection:
\begin{equation}
\mathbf{U}^{\,l}_{t}
=
\mathrm{CBR}
\left(
\mathbf{M}^{\,l}_{t}
\odot
\mathbf{V}^{\,l}_{t}
\right),
\qquad
\mathrm{FCA}
\left(
\mathbf{g}^{\,l},\mathbf{f}^{\,l}_{t}
\right)
=
\mathbf{f}^{\,l}_{t}
+
\mathbf{U}^{\,l}_{t} 
\end{equation}
where $\odot$ denotes element-wise modulation.

\subsection{Multi-Scale Cross-Transformer Reconstruction (MSCTR)}

\begin{figure}[!t]
    \centering
    \includegraphics[width=\linewidth]{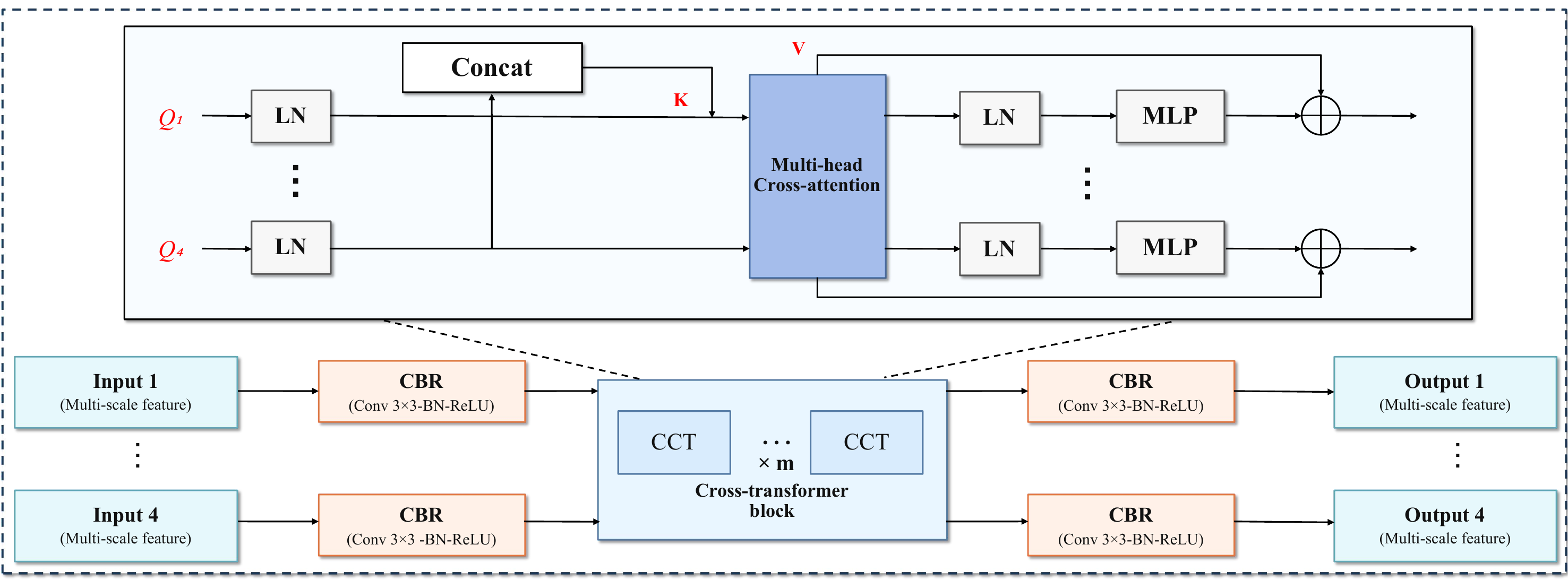}
    \caption{MSCTR module for decoder-stage multi-scale feature reconstruction. Encoder features and AFG features are fused into multi-scale skip tokens, processed by stacked CCT blocks, and injected into the decoder through residual connections. }
    \label{fig:msctr_framework}
\end{figure}

After encoder-stage anatomical-frequency modulation, MSCTR reconstructs decoder skip features to mitigate feature drift across diffusion steps. 
Instead of directly forwarding encoder features to the decoder, MSCTR first aggregates multi-scale AFG-refined features within the same representation stage. 
This design enables interaction across multiple resolutions in a unified feature space, allowing the model to jointly capture global structural context and fine-grained anatomical details. 
The aggregated multi-scale features are then processed by Cross-Transformer (CCT) blocks, which exchange information across resolutions to enhance structural consistency and stabilize decoder reconstruction.
This design reduces noisy feature propagation and improves decoder-stage structural consistency. 
Fig.~\ref{fig:msctr_framework} illustrates the module structure.

At diffusion step $t$, MSCTR receives the encoder feature of the $l$-th layer $\mathbf{f}^{\,l}_{t}$ and the corresponding AFG feature $\mathbf{f}^{\,l,\mathrm{out}}_{t}$ at each scale $n \in \{1,2,3,4\}$.

Within the same layer, all multi-scale features are first aggregated by fusing the AFG-refined representations through channel-wise concatenation, and then uniformly processed using a shared CBR block to obtain refined skip representations:
\begin{equation}
\mathbf{s}^{\,l,n}_{t}
=
\mathrm{CBR}
\left(
\operatorname{Concat}
\left(
\mathbf{f}^{\,l}_{t},
\mathbf{f}^{\,l,\mathrm{out}}_{t}
\right)
\right),
\qquad
n \in \{1,2,3,4\}.
\end{equation}
where $\operatorname{Concat}(\cdot)$ denotes channel-wise concatenation.

To enable cross-scale feature reconstruction, all scale-specific skip representations are flattened into token sequences and used jointly in a CCT block. 
For the $l$-th layer and the $n$-th scale, the query is generated from its own scale-specific skip representation, while the keys and values are aggregated across all scales:
\begin{equation}
\begin{aligned}
\widetilde{\mathbf{Q}}^{\,l,n}_{t}
&=
\mathrm{MLP}\big(\mathrm{LN}(\mathbf{s}^{\,l,n}_{t})\big),\\
\widetilde{\mathbf{K}}^{\,l}_{t}
&=
\operatorname{Concat}_{n=1}^{4}
\mathrm{MLP}\big(\mathrm{LN}(\mathbf{s}^{\,l,n}_{t})\big),\\
\widetilde{\mathbf{V}}^{\,l}_{t}
&=
\operatorname{Concat}_{n=1}^{4}
\mathrm{MLP}\big(\mathrm{LN}(\mathbf{s}^{\,l,n}_{t})\big)
\end{aligned}
\end{equation}
The cross-scale attention output for scale $n$ is computed as
\begin{equation}
\hat{\mathbf{s}}^{\,l}_{t}
=
\mathbf{s}^{\,l,n}_{t}
+
\mathrm{MSCA}
\left(
\widetilde{\mathbf{Q}}^{\,l,n}_{t},
\widetilde{\mathbf{K}}^{\,l}_{t},
\widetilde{\mathbf{V}}^{\,l}_{t}
\right) 
\end{equation}
where $\mathrm{MSCA}(\cdot)$ denotes multi-head cross-attention. 
A feed-forward MLP block is then applied with a residual connection at $l$-th layer :
\begin{equation}
\tilde{\mathbf{s}}^{\,l}_{t}
=
\hat{\mathbf{s}}^{\,l}_{t}
+
\mathrm{MLP}
\left(
\mathrm{LN}
\left(
\hat{\mathbf{s}}^{\,l}_{t}
\right)
\right) 
\end{equation}
Stacking this operation for $m$ times forms the CCT blocks shown in Fig.~\ref{fig:msctr_framework}.

Finally, the reconstructed multi-scale skip feature is reshaped back to the spatial feature map and injected into the decoder through residual fusion:
\begin{equation}
\mathbf{d}^{\,l,\mathrm{out}}_{t}
=
\mathbf{d}^{\,l}_{t}
+
\mathrm{CBR}
\left(
\tilde{\mathbf{s}}^{\,l}_{t}
\right) 
\end{equation}
where $\mathbf{d}^{\,l}_{t}$ and $\mathbf{d}^{\,l,\mathrm{out}}_{t}$ denote the decoder feature before and after MSCTR reconstruction, respectively.

\subsection{Frequency-Contrastive Hard Mining (FCHM)}

After MSCTR reconstructs decoder features, we further regularize the denoised HD PET estimate in the frequency domain during training. 
FCHM encourages the predicted HD PET spectrum to approach its paired HD PET target while being separated from hard negative LD spectra within the mini-batch.

The denoising network takes the paired LD PET/CT input $(\mathbf{y}_{\mathrm{LD}},\mathbf{z})$ as the conditional input and predicts the denoised HD PET image $\hat{\mathbf{x}}_{\mathrm{HD}}$, which serves as the anchor. 
The positive sample is the paired HD PET image $\mathbf{x}_{\mathrm{HD}}$. 
The negative samples are formed by the paired LD PET image $\mathbf{y}_{\mathrm{LD}}$ and other LD PET images $\{\mathbf{y}_{\mathrm{LD}}^{other}\}$ from the same mini-batch:
\begin{equation}
\mathcal{N}_{\mathrm{LD}}
=
\{\mathbf{y}_{\mathrm{LD}}\}
\cup
\{\mathbf{y}_{\mathrm{LD}}^{other}\} 
\end{equation}
where $\mathcal{N}_{\mathrm{LD}}$ denotes the set of all LD PET negative samples.

Thus, FCHM optimizes the anchor prediction $\hat{\mathbf{x}}_{\mathrm{HD}}$ by pulling its frequency spectrum toward the paired target $\mathbf{x}_{\mathrm{HD}}$ and pushing it away from samples in $\mathcal{N}_{\mathrm{LD}}$ within the same mini-batch.

First, the anchor, positive target, and LD negative samples are transformed into the frequency domain via two-dimensional FFT:
\begin{equation}
\begin{aligned}
\mathcal{F}(\hat{\mathbf{x}}_{\mathrm{HD}})
&=
\mathrm{FFT}
\left(
\hat{\mathbf{x}}_{\mathrm{HD}}
\right),
&
\mathcal{F}(\mathbf{x}_{\mathrm{HD}})
&=
\mathrm{FFT}
\left(
\mathbf{x}_{\mathrm{HD}}
\right),
\\
\mathcal{F}(\mathbf{y}^j)
&=
\mathrm{FFT}
\left(
\mathbf{y}^j
\right),
&
\mathbf{y}^j
&\in
\mathcal{N}_{\mathrm{LD}} 
\end{aligned}
\end{equation}
where $\mathbf{y}^j \in \mathcal{N}_{\mathrm{LD}}$ denotes the $j$-th negative sample in the mini-batch.

Then, we compute the L2  distance between the anchor prediction $\hat{\mathbf{x}}_{\mathrm{HD}}$ and its positive target $\mathbf{x}_{\mathrm{HD}}$ in the frequency domain:
\begin{equation}
d_{ap}
=
\left\|
\left|
\mathcal{F}
\left(
\hat{\mathbf{x}}_{\mathrm{HD}}
\right)
\right|
-
\left|
\mathcal{F}
\left(
\mathbf{x}_{\mathrm{HD}}
\right)
\right|
\right\|_{2}
\end{equation}

Similarly, the L2 spectral distance between the anchor and each LD negative sample $\mathbf{y}^{j}\in\mathcal{N}_{\mathrm{LD}}$ is defined as:
\begin{equation}
d_{an}^{j}
=
\left\|
\left|
\mathcal{F}
\left(
\hat{\mathbf{x}}_{\mathrm{HD}}
\right)
\right|
-
\left|
\mathcal{F}
\left(
\mathbf{y}^{j}
\right)
\right|
\right\|_{2}
\end{equation}

Next, an online hard example mining (OHEM) strategy is used to select the Top-$k$ negatives with the smallest $d_{an}$. 
Let $\mathcal{H}_{k}$ denote the selected Top-$k$ hard negative set, and the frequency domain contrastive loss is defined as
\begin{equation}
L_{\mathrm{FCHM}}
=
\frac{d_{ap}}
{
\frac{1}{|H_k|}
\sum_{j\in H_k}
d_{an}^{j}
+
\delta
} 
\end{equation}
where $\delta$ is a small constant for numerical stability. 
Unless otherwise specified, we set $\delta=10^{-6}$ in all experiments.

Following the prediction target defined above, the spatial domain reconstruction loss is applied to the $\hat{\mathbf{x}}_{\mathrm{HD}}$:
\begin{equation}
\mathcal{L}_{1}
=
\mathbb{E}_{\mathbf{x}_{\mathrm{HD}},\mathbf{y}_{\mathrm{LD}},\mathbf{z},t,\boldsymbol{\epsilon}}
\left[
\left\|
\hat{\mathbf{x}}_{\mathrm{HD}}
-
\mathbf{x}_{\mathrm{HD}}
\right\|_{1}
\right] 
\end{equation}
where $\hat{\mathbf{x}}_{\mathrm{HD}}=f_{\theta}(\mathbf{x}_t,\mathbf{y}_{\mathrm{LD}},\mathbf{z},t)$ and $\mathbf{x}_t$ is sampled from the forward diffusion process. 
The overall training objective combines this spatial domain reconstruction loss and the proposed frequency domain contrastive regularization:
\begin{equation}
\mathcal{L}_{\mathrm{total}}
=
\mathcal{L}_{1}
+
\lambda_{\mathrm{FCHM}}
\mathcal{L}_{\mathrm{FCHM}} 
\label{eq:total_loss}
\end{equation}

This joint objective preserves PET intensity fidelity in the spatial domain while improving frequency domain consistency for stable denoising.

\section{Experiments}

\subsection{Experimental Setup}

\subsubsection{Datasets}
We evaluate AnF-DiffPET on four PET/CT datasets covering different tracers, anatomical regions, and disease contexts, including Head and Neck Tumor (HECKTOR) PET/CT~\cite{dataset}, Prostate-Specific Membrane Antigen PET/CT Lesions (PSMA)~\cite{psma_pet_ct_lesions}, $^{18}$F-Sodium Fluoride Prostate PET/CT (NaF)~\cite{naf_prostate_tcia}, and Reference Image Database to Evaluate Therapy Response Lung PET/CT (RIDER)~\cite{rider_lung_pet_ct_tcia}.
These datasets provide a diverse multi-cohort evaluation setting for assessing LD PET denoising performance under different uptake distributions and anatomical conditions.
For each dataset, the training and testing partitions are defined at the PET/CT study level before slice extraction, so that slices from the same scan are not shared across partitions.
PET/CT volumes are then preprocessed and converted into paired axial samples within each partition.
During this process, slices dominated by background or containing very limited PET uptake and anatomical information are excluded, while informative slices are retained from different scans and anatomical levels to preserve the diversity of PET appearances.

HECKTOR consists of coregistered PET and CT image pairs from head and neck cancer imaging.
After preprocessing, we obtain 5{,}000 paired PET/CT samples, including 4{,}000 samples for training and 1{,}000 samples for testing.

PSMA contains prostate cancer related PET/CT images acquired with prostate-specific membrane antigen targeted radiotracers.
This cohort is used to evaluate the recovery of focal molecular uptake patterns in prostate cancer imaging.
After preprocessing, we retain 5{,}000 paired PET/CT samples covering different scans and uptake patterns, with 4{,}000 samples for training and 1{,}000 samples for testing.

NaF contains $^{18}$F-sodium fluoride PET/CT scans from patients with prostate cancer.
Since NaF PET reflects bone metabolism, this dataset provides a distinct evaluation scenario focused on skeletal uptake patterns and high contrast bone related structures.
After preprocessing, 4{,}545 paired samples are retained, including 3{,}629 samples for training and 916 samples for testing.

RIDER contains lung cancer PET/CT scans and is used to evaluate thoracic PET/CT denoising.
Compared with the prostate cohorts, RIDER includes different anatomical structures, uptake distributions, and noise characteristics.
After preprocessing, 4{,}539 paired samples are obtained, including 3{,}638 samples for training and 901 samples for testing.

\subsubsection{PET/CT preprocessing}

For each dataset, we start from reconstructed three-dimensional PET volumes and the corresponding CT volumes from the same PET/CT study.
The reconstructed HD PET volume is used as the reference target, and the LD PET volume is generated from it during preprocessing.
Paired axial samples are then extracted from the LD PET volume, the HD PET volume, and the corresponding CT volume.

PET images are represented on the standardized uptake value (SUV) intensity scale for quantitative evaluation.
For network training, PET intensities are clipped to a clinically relevant range and normalized to $[0,1]$.
\begin{equation}
\mathbf{x}_{\mathrm{HD}}^{\mathbf{v}}
=
\frac{
\operatorname{clip}
\left(
\mathbf{x}_{\mathrm{SUV}},0,20
\right)
}{20} 
\label{eq:pet_suv_norm}
\end{equation}
where $\mathbf{x}_{\mathrm{SUV}}$ denotes the reconstructed HD PET volume in SUV units, and $\mathbf{x}_{\mathrm{HD}}^{\mathbf{v}}$ denotes the normalized HD reference volume.
The original SUV scaling is retained for SUV based quantitative evaluation.
CT intensities are clipped using a Hounsfield unit window of $[-1000,1000]$ HU and then normalized to $[0,1]$.

\subsubsection{Poisson based LD PET generation}

To construct paired LD and HD PET samples, we simulate count limited PET degradation in the reconstructed volume domain.
This setting follows the common LD PET denoising formulation, where reduced tracer dose or shortened acquisition time leads to fewer detected photon counts, stronger noise, and lower signal-to-noise ratio.
We approximate the low-count process on reconstructed PET volumes.

Let $\mathbf{x}_{\mathrm{HD}}^{\mathbf{v}}\in[0,1]^{H\times W\times S}$ denote the normalized HD PET volume.
We first convert the normalized PET intensity into a pseudo count representation.
\begin{equation}
C_{\mathrm{HD}}
=
\kappa \mathbf{x}_{\mathrm{HD}}^{\mathbf{v}}
\label{eq:pseudo_count_hd}
\end{equation}
where $\kappa=1000$ is a fixed global pseudo count scaling constant controlling the nominal count level of the reference PET volume, which represents the image domain noise magnitude.

To simulate an LD acquisition, we sample a reduced count volume using Poisson statistics.
\begin{equation}
C_{\mathrm{LD}}(p)
\sim
\mathrm{Poisson}
\left(
\rho C_{\mathrm{HD}}(p)
\right),
\qquad
0<\rho<1 
\label{eq:poisson_ld_sampling}
\end{equation}
where $p$ indexes voxels and $\rho$ denotes the relative count level of the simulated LD PET volume.
A smaller $\rho$ corresponds to fewer effective counts and stronger LD noise.
In our implementation, we set $\rho=0.25$, corresponding to one quarter of the nominal count level.

The sampled LD pseudo count volume is then converted back to the normalized PET intensity scale by compensating for the expected count reduction.
\begin{equation}
\mathbf{x}_{\mathrm{LD}}^{\mathbf{v}}
=
\operatorname{clip}
\left(
\frac{
C_{\mathrm{LD}}
}{
\rho\kappa
},
0,1
\right) 
\label{eq:ld_pet_generation}
\end{equation}
where $\mathbf{x}_{\mathrm{LD}}^{\mathbf{v}}$ denotes the simulated LD PET volume.
Under this compensation, the expected PET uptake intensity is preserved, while count dependent Poisson noise is increased.

After preprocessing, paired axial samples are extracted from $\mathbf{x}_{\mathrm{LD}}^{\mathbf{v}}$, $\mathbf{x}_{\mathrm{HD}}^{\mathbf{v}}$, and the normalized CT volume.
\begin{equation}
\begin{aligned}
\mathbf{x}_{\mathrm{LD}}
&=
\mathcal{R}_{128}
\left(
\mathbf{x}_{\mathrm{LD}}^{\mathbf{v}}
\right),
\\
\mathbf{x}_{\mathrm{HD}}
&=
\mathcal{R}_{128}
\left(
\mathbf{x}_{\mathrm{HD}}^{\mathbf{v}}
\right) 
\end{aligned}
\end{equation}
\begin{equation}
\mathbf{z}
=
\mathcal{R}_{128}
\left(
\mathrm{Z}^{\mathbf{v}}
\right) 
\end{equation}
where $\mathbf{Z}$ denotes the normalized CT volume, and $\mathcal{R}_{128}(\cdot)$ denotes resizing or cropping to $128\times128$.
Each training sample is organized as
\begin{equation}
\left(
\mathbf{x}_{\mathrm{LD}},
\mathbf{z},
\mathbf{x}_{\mathrm{HD}}
\right) 
\end{equation}
where $\mathbf{x}_{\mathrm{LD}}$ is the LD PET input, $\mathbf{z}$ is the paired CT image, and $\mathbf{x}_{\mathrm{HD}}$ is the raw image.
For fair comparison, the same pre-generated LD PET volumes are used for all compared methods.
This preprocessing provides paired LD and HD PET/CT samples for supervised PET denoising.
\begin{figure}[!t]
    \centering
    \includegraphics[width=\linewidth]{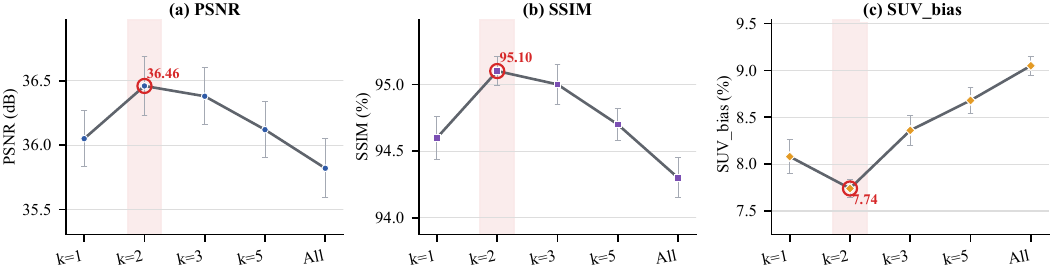}
    \caption{FCHM hard-negative sensitivity on HECKTOR. Three panels show PSNR, SSIM, and SUV\_bias as functions of the hard-negative setting $k$.}
    \label{fig:k_sensitivity_line}
\end{figure}

\begin{table}[!t]
\centering
\caption{Sensitivity of FCHM to the loss weight $\lambda_{\mathrm{FCHM}}$ on HECKTOR. Values are reported as mean $\pm$ standard deviation where applicable, and SSIM and SUV\_bias are reported in percentage. $E_l$, $E_m$, and $E_h$ denote low-, middle-, and high-frequency spectral reconstruction errors, respectively, defined over normalized radial frequency ranges of $0$--$30\%$, $30\%$--$70\%$, and $70\%$--$100\%$ in the centered Fourier domain. Bold values indicate the best result for each metric.}
\label{tab:lambda_fchm_sensitivity}
\scriptsize
\setlength{\tabcolsep}{3pt}
\renewcommand{\arraystretch}{1.15}
\resizebox{\linewidth}{!}{%
\begin{tabular}{lcccccc}
\toprule
$\lambda_{\mathrm{FCHM}}$
& PSNR~(dB)~$\uparrow$
& SSIM~(\%)~$\uparrow$
& SUV\_bias~(\%)~$\downarrow$
& $E_l~\downarrow$
& $E_m~\downarrow$
& $E_h~\downarrow$ \\
\midrule
0
& $35.93\pm0.24$ & $94.60\pm0.14$ & $8.05\pm0.16$
& 0.276 & 0.842 & 1.238 \\
0.10
& $36.12\pm0.23$ & $94.76\pm0.13$ & $7.97\pm0.15$
& 0.268 & 0.834 & 1.184 \\
0.25
& $36.27\pm0.23$ & $94.91\pm0.12$ & $7.88\pm0.13$
& 0.259 & 0.826 & 1.126 \\
0.50
& $36.39\pm0.23$ & $95.02\pm0.11$ & $7.80\pm0.11$
& 0.253 & \textbf{0.809} & 1.079 \\
1.00
& $\mathbf{36.46\pm0.23}$ & $\mathbf{95.10\pm0.11}$ & $\mathbf{7.74\pm0.10}$
& \textbf{0.249} & 0.813 & \textbf{1.058} \\
2.00
& $36.21\pm0.25$ & $94.92\pm0.13$ & $7.92\pm0.13$
& 0.265 & 0.824 & 1.101 \\
5.00
& $35.68\pm0.28$ & $94.31\pm0.17$ & $8.46\pm0.18$
& 0.303 & 0.876 & 1.254 \\
\bottomrule
\end{tabular}
}
\end{table}

\subsubsection{Evaluation metrics}
Denoising quality is evaluated using peak signal-to-noise ratio (PSNR), structural similarity index (SSIM), and absolute $\mathrm{SUV}_{\mathrm{mean}}$ bias (SUV\_bias).
Throughout evaluation, the raw image is used as the ground truth.
Let $\hat{\mathbf{x}}_{\mathrm{HD}}$ denote the denoised PET image on the evaluation grid $\Omega$, with dynamic range $L$.
The PSNR is computed as
\begin{equation}
\mathrm{PSNR}
=
10\log_{10}
\left(
\frac{
L^2
}{
\frac{1}{|\Omega|}
\sum_{p\in\Omega}
\left(
\hat{\mathbf{x}}_{\mathrm{HD}}(p)-\mathbf{x}_{\mathrm{HD}}(p)
\right)^2
}
\right) 
\end{equation}
PSNR reflects residual error after logarithmic normalization by the image dynamic range.
SSIM is computed from local intensity statistics between the denoised PET image and the raw image as
\begin{equation}
\mathrm{SSIM}
=
\frac{
\left(2\mu_{\mathrm{R}}\mu_{\mathrm{HD}}+C_1\right)
\left(2\sigma_{\mathrm{R,HD}}+C_2\right)
}{
\left(\mu_{\mathrm{R}}^2+\mu_{\mathrm{HD}}^2+C_1\right)
\left(\sigma_{\mathrm{R}}^2+\sigma_{\mathrm{HD}}^2+C_2\right)
} 
\end{equation}
where $\mu_{\mathrm{R}}$ and $\mu_{\mathrm{HD}}$ are local means, $\sigma_{\mathrm{R}}^2$ and $\sigma_{\mathrm{HD}}^2$ are local variances, $\sigma_{\mathrm{R,HD}}$ is the local covariance, and $C_1$ and $C_2$ are stabilization constants.

To evaluate PET quantitative fidelity, we further compute SUV\_bias within the PET foreground mask.
The foreground mask $\Omega$ is derived from the ground truth PET image and is fixed for both the denoised and reference images during evaluation.
For SUV based evaluation, PET images are measured on the SUV intensity scale. When intensity normalization is applied for network training, the corresponding inverse normalization is used before computing SUV statistics.

For $q\in\{\hat{\mathbf{x}}_{\mathrm{HD}},\mathbf{x}_\mathrm{HD}\}$, the foreground $\mathrm{SUV}_{\mathrm{mean}}$ is computed as
\begin{equation}
s_q
=
\mathrm{SUV}_{\mathrm{mean}}(\mathbf{x}_q;\Omega)
=
\frac{1}{|\Omega|}
\sum_{p\in\Omega}
\mathbf{x}_q(p) 
\end{equation}
where $\hat{\mathbf{x}}_{\mathrm{HD}}$ and $\mathbf{x}_{\mathrm{HD}}$ denote the denoised PET image and the raw image, respectively, and $p$ indexes pixels inside the foreground mask $\Omega$.
The SUV\_bias is then computed as
\begin{equation}
\mathrm{SUV\_bias}
=
100
\left|
\frac{
s_{\hat{\mathbf{x}}_{\mathrm{HD}}}-s_{\mathbf{x}_\mathrm{HD}}
}{
s_{\mathbf{x}_\mathrm{HD}}+\eta
}
\right| 
\label{eq:suvmean_bias}
\end{equation}
where $\eta=10^{-8}$ is used for numerical stability.

\begin{table*}[!htbp]
\centering
\caption{Quantitative comparisons on the HECKTOR, PSMA, NaF, and RIDER datasets. PSNR in dB, SSIM and SUV\_bias in \%.}
\label{tab:main_results}
\begin{threeparttable}
\setlength{\tabcolsep}{4pt}
\renewcommand{\arraystretch}{1.15}
\resizebox{\textwidth}{!}{%
\begin{tabular}{l*{12}{c}}
\toprule
\multirow{2}{*}{\textbf{Method}}
& \multicolumn{3}{c}{\textbf{HECKTOR}}
& \multicolumn{3}{c}{\textbf{PSMA}}
& \multicolumn{3}{c}{\textbf{NaF}}
& \multicolumn{3}{c}{\textbf{RIDER}} \\
\cmidrule(lr){2-4} \cmidrule(lr){5-7} \cmidrule(lr){8-10} \cmidrule(lr){11-13}
& PSNR$\uparrow$ & SSIM$\uparrow$ & Bias$\downarrow$
& PSNR$\uparrow$ & SSIM$\uparrow$ & Bias$\downarrow$
& PSNR$\uparrow$ & SSIM$\uparrow$ & Bias$\downarrow$
& PSNR$\uparrow$ & SSIM$\uparrow$ & Bias$\downarrow$ \\
\midrule
ESRGAN
& $29.03\pm0.24^{*}$ & $82.40\pm0.14^{*}$ & $19.84\pm0.16^{*}$
& $28.53\pm0.22^{*}$ & $89.60\pm0.12^{*}$ & $22.18\pm0.12^{*}$
& $29.12\pm0.23^{*}$ & $88.70\pm0.13^{*}$ & $20.06\pm0.15^{*}$
& $27.84\pm0.23^{*}$ & $87.20\pm0.14^{*}$ & $23.48\pm0.11^{*}$ \\
VDSR
& $31.08\pm0.24^{*}$ & $87.50\pm0.13^{*}$ & $18.52\pm0.11^{*}$
& $30.06\pm0.23^{*}$ & $92.60\pm0.13^{*}$ & $21.06\pm0.13^{*}$
& $30.48\pm0.23^{*}$ & $91.80\pm0.13^{*}$ & $19.43\pm0.11^{*}$
& $29.36\pm0.23^{*}$ & $90.90\pm0.12^{*}$ & $21.76\pm0.16^{*}$ \\
EDSR
& $31.25\pm0.23^{*}$ & $87.60\pm0.17^{*}$ & $17.96\pm0.18^{*}$
& $30.94\pm0.23^{*}$ & $93.60\pm0.19^{*}$ & $18.34\pm0.17^{*}$
& $31.07\pm0.22^{*}$ & $92.90\pm0.17^{*}$ & $17.87\pm0.12^{*}$
& $30.18\pm0.23^{*}$ & $92.20\pm0.17^{*}$ & $18.09\pm0.14^{*}$ \\
CDM-GAN
& $32.03\pm0.26^{*}$ & $88.10\pm0.12^{*}$ & $11.21\pm0.17^{*}$
& $31.15\pm0.25^{*}$ & $93.40\pm0.12^{*}$ & $11.48\pm0.12^{*}$
& $31.82\pm0.25^{*}$ & $93.70\pm0.12^{*}$ & $11.34\pm0.14^{*}$
& $30.87\pm0.25^{*}$ & $93.00\pm0.12^{*}$ & $11.68\pm0.16^{*}$ \\
SwinIR
& $32.13\pm0.25^{*}$ & $89.40\pm0.18^{*}$ & $10.47\pm0.11^{*}$
& $32.79\pm0.25^{*}$ & $93.70\pm0.21^{*}$ & $10.72\pm0.14^{*}$
& $32.18\pm0.25^{*}$ & $94.10\pm0.20^{*}$ & $10.51\pm0.16^{*}$
& $31.54\pm0.25^{*}$ & $93.60\pm0.19^{*}$ & $10.74\pm0.18^{*}$ \\
I2SB
& $32.07\pm0.25^{*}$ & $90.50\pm0.16^{*}$ & $9.93\pm0.16^{*}$
& $30.27\pm0.24^{*}$ & $93.10\pm0.18^{*}$ & $10.15\pm0.09^{*}$
& $31.76\pm0.24^{*}$ & $94.40\pm0.17^{*}$ & $9.84\pm0.11^{*}$
& $31.22\pm0.24^{*}$ & $93.90\pm0.17^{*}$ & $10.12\pm0.08^{*}$ \\
RDDM
& $33.30\pm0.24^{*}$ & $91.60\pm0.15^{*}$ & $8.28\pm0.14^{*}$
& $32.54\pm0.24^{*}$ & $96.40\pm0.15^{*}$ & $8.54\pm0.16^{*}$
& $32.86\pm0.23^{*}$ & $96.20\pm0.14^{*}$ & $8.35\pm0.17^{*}$
& $32.37\pm0.24^{*}$ & $95.80\pm0.15^{*}$ & $8.71\pm0.10^{*}$ \\
\midrule
\textbf{Ours}
& $\mathbf{36.46\pm0.23}$ & $\mathbf{95.10\pm0.11}$ & $\mathbf{7.74\pm0.10}$
& $\mathbf{33.49\pm0.22}$ & $\mathbf{97.80\pm0.11}$ & $\mathbf{8.07\pm0.11}$
& $\mathbf{34.21\pm0.22}$ & $\mathbf{97.40\pm0.11}$ & $\mathbf{7.92\pm0.12}$
& $\mathbf{33.76\pm0.22}$ & $\mathbf{97.10\pm0.10}$ & $\mathbf{8.28\pm0.12}$ \\
\bottomrule
\end{tabular}%
}
\begin{tablenotes}
\footnotesize
\item[*] denotes statistical significance (paired two-sided $t$-test, $p<0.05$).
\end{tablenotes}
\end{threeparttable}
\label{compar}
\end{table*}

\begin{figure*}[!tbp]
    \centering
    \includegraphics[width=0.8\textwidth]{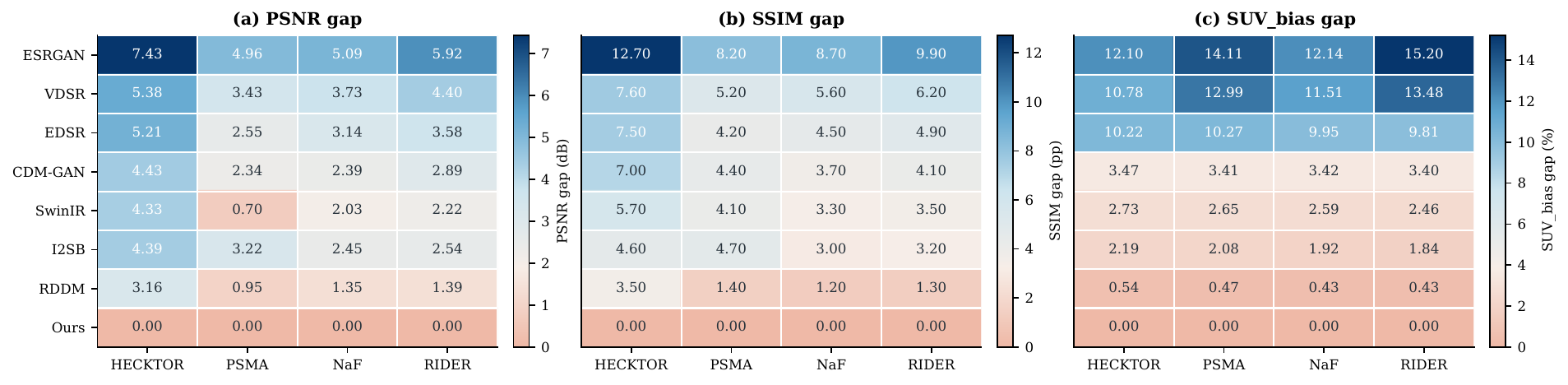}
    \caption{Annotated heatmaps of performance gaps between AnF-DiffPET and compared methods across four PET/CT datasets. Gaps are computed as AnF-DiffPET minus the compared method for PSNR and SSIM, and as the compared method minus AnF-DiffPET for SUV\_bias; larger values indicate larger advantages of AnF-DiffPET.}
    \label{fig:quantitative_metric_trends}
\end{figure*}

\begin{figure*}[!t]
    \centering
    \includegraphics[width=0.8\textwidth]{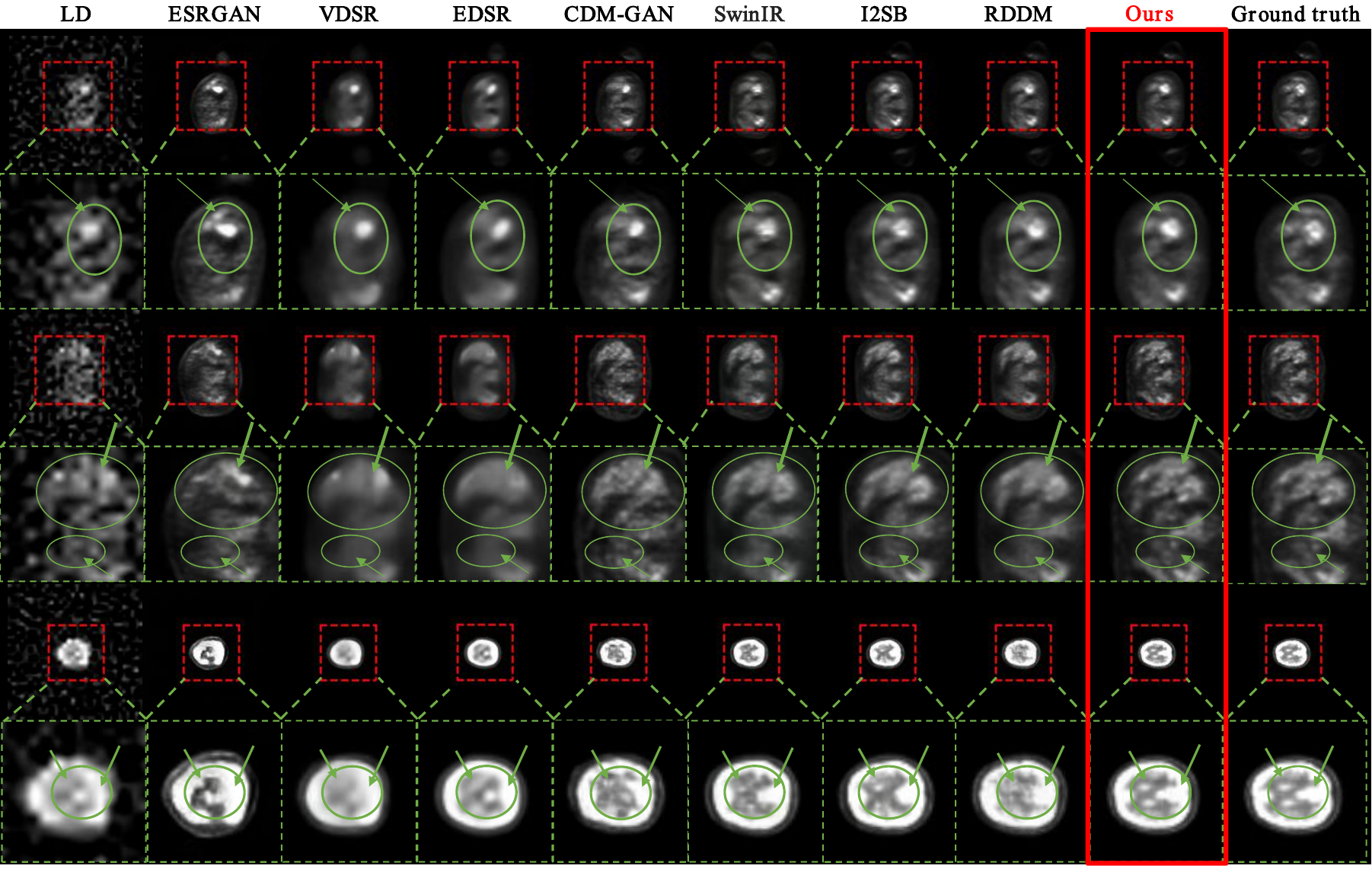}
    \caption{Representative HECKTOR PET denoising case. Columns show the LD PET input, outputs from compared methods, AnF-DiffPET, and the ground truth HD PET image. For each case, the first row shows the full PET slice, and the second row shows the ROI zoom-in  views corresponding to the dashed box in the first row. Green arrows and circles highlight representative uptake structures and local anatomical details for visual comparison.}
    \label{fig:sota}
\end{figure*}

\begin{figure*}[!tbp]
    \centering
    \setlength{\abovecaptionskip}{2pt}
    \setlength{\belowcaptionskip}{0pt}
    \includegraphics[width=0.8\textwidth]{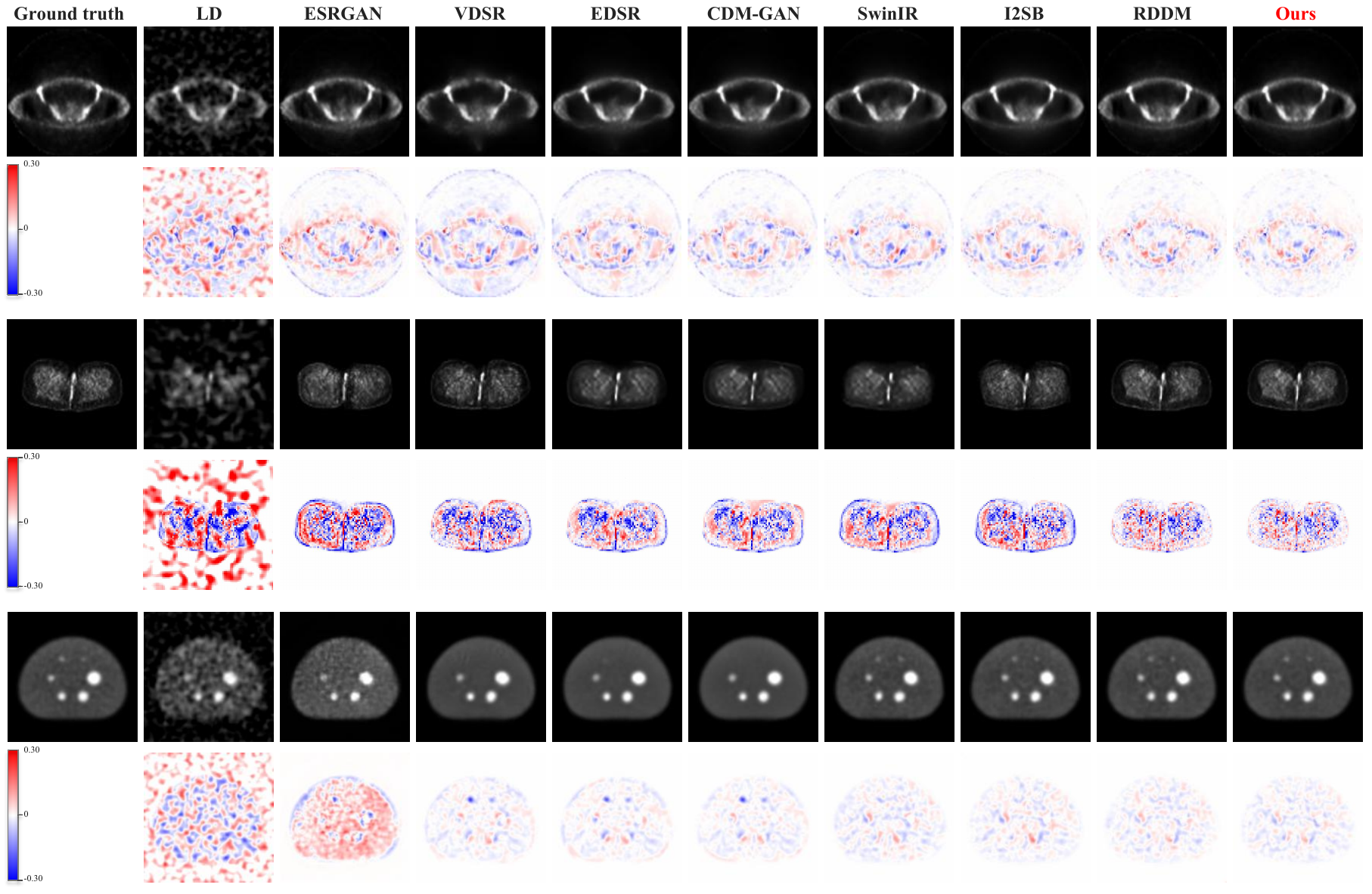}
    \caption{Representative PET/CT denoising cases on NaF, PSMA, and RIDER. From top to bottom, the three case groups correspond to NaF, PSMA, and RIDER, respectively. Columns show the ground truth HD PET image, LD PET input, outputs from competing methods, and AnF-DiffPET. For each case group, the upper row presents the denoised PET image, whereas the lower row shows the signed residual error map computed as prediction minus ground truth. Red regions indicate overestimated uptake, whereas blue regions indicate underestimated uptake.}
    \label{fig:external_cases}
\end{figure*}
\subsubsection{Implementation details}
All methods are trained on an NVIDIA RTX 5090 GPU using the Adam optimizer with an initial learning rate of $2\times10^{-4}$.
The diffusion process uses a 1{,}000-step linear noise schedule.
During training, the noisy target image $\mathbf{x}_t$ is sampled from the forward diffusion process applied to the raw image, and the denoising network is conditioned on the paired LD PET/CT input.
Under the $\mathbf{x}_{\mathrm{HD}}$ prediction parameterization, the network output is the denoised HD PET estimate used by the $\mathcal{L}_{1}$ and FCHM losses.
During inference, the reverse sampling process is initialized from Gaussian noise and progressively reconstructs the denoised HD PET estimate under LD PET/CT conditioning.
The denominator term in FCHM is computed as the average spectral distance over the selected top-$k$ hard negatives after online hard example mining.


\subsection{Hyperparameter selection}

We further examine the sensitivity of FCHM to two key hyperparameters, the number of hard negatives $k$ and the loss weight $\lambda_{\mathrm{FCHM}}$.
The hard negative number $k$ controls the selectivity of frequency domain negative mining, while $\lambda_{\mathrm{FCHM}}$ balances the spatial domain reconstruction objective and the frequency domain contrastive regularization.
These experiments are conducted on the HECKTOR dataset with all other training settings fixed, aiming to justify the default choices used in the final method.

\subsubsection{Hard Negative Sensitivity in FCHM}
FCHM selects the top-$k$ hardest negative samples in each mini-batch according to their frequency domain distance from the anchor prediction.
A smaller $k$ imposes a more selective hard negative constraint but may increase sensitivity to outlier negatives, whereas a larger $k$ introduces more ordinary negatives and may dilute the mining effect.
We therefore vary $k$ while keeping all other settings unchanged, evaluating $k=1$, $k=2$, $k=3$, $k=5$, and all negatives in the batch.
We report PSNR, SSIM and SUV\_bias together to evaluate image fidelity, structural similarity, and quantitative fidelity.
As shown in Fig.~\ref{fig:k_sensitivity_line}, $k=2$ achieves the best overall performance, with the highest PSNR and SSIM and the lowest SUV\_bias.
When $k=1$, FCHM uses only the single hardest negative sample, which provides a strong frequency domain constraint but may make training more sensitive to outlier negatives.
Increasing $k$ to 3 yields comparable performance, suggesting that a moderate number of hard negatives can still provide stable supervision.
However, further increasing $k$ to 5 or using all negatives leads to a gradual performance drop, indicating that excessive inclusion of ordinary negatives weakens the hard mining effect and makes the objective closer to a generic frequency domain contrastive loss.
Therefore, we use $k=2$ as the default FCHM setting.

\subsubsection{Sensitivity to the FCHM Loss Weight}

We further analyze the effect of the FCHM loss weight $\lambda_{\mathrm{FCHM}}$.
The overall training objective is defined in Eq.~\eqref{eq:total_loss}.

Because $\lambda_{\mathrm{FCHM}}$ controls the strength of frequency domain contrastive supervision, its selection can affect not only image domain fidelity but also the recovery of spectral details across different frequency bands.

For the design sensitivity analysis of FCHM, we additionally report band wise spectral reconstruction errors to evaluate frequency domain consistency.
Let $\Omega_l$, $\Omega_m$, and $\Omega_h$ denote the low, middle, and high frequency regions in the centered Fourier domain, respectively.
For a frequency band $b\in\{l,m,h\}$, the spectral reconstruction error is computed as
\begin{equation}
\begin{aligned}
E_b
&=
\frac{1}{|\Omega_b|}
\left\|
\left(
\left|\mathcal{F}(\hat{\mathbf{x}}_{\mathrm{HD}})\right|
-
\left|\mathcal{F}(\mathbf{x}_{\mathrm{HD}})\right|
\right)_{\Omega_b}
\right\|_2,
\\
&\hspace{2.0em} b\in\{l,m,h\} 
\end{aligned}
\end{equation}
where $\mathcal{F}(\cdot)$ denotes the two dimensional FFT and $|\cdot|$ denotes the magnitude spectrum.
Thus, $E_l$, $E_m$, and $E_h$ measure low, middle, and high frequency spectral discrepancies, respectively. Lower values indicate better frequency domain consistency with the ra w.

To isolate the effect of $\lambda_{\mathrm{FCHM}}$, we keep all other settings fixed, including $k=2$ and $\delta=10^{-6}$ in FCHM, and vary $\lambda_{\mathrm{FCHM}}$ from 0 to 5.
Here, $\lambda_{\mathrm{FCHM}}=0$ corresponds to disabling the FCHM regularization.

Table~\ref{tab:lambda_fchm_sensitivity} shows that $\lambda_{\mathrm{FCHM}}=1.0$ provides the best overall performance. At this setting, The standard deviations are also among the smallest across metrics, indicating stable improvements across test samples.
When $\lambda_{\mathrm{FCHM}}=0$, the FCHM term is removed and the method relies only on spatial domain reconstruction, leading to larger SUV\_bias and larger spectral errors.
As $\lambda_{\mathrm{FCHM}}$ increases from 0 to 1.0, PSNR and SSIM improve steadily, while SUV\_bias and the low and high frequency spectral errors decrease.
This indicates that FCHM provides complementary frequency domain supervision beyond the pixel wise reconstruction loss. When $\lambda_{\mathrm{FCHM}}$ becomes too large, performance starts to degrade.
For example, increasing $\lambda_{\mathrm{FCHM}}$ to 2.0 or 5.0 reduces PSNR and SSIM and increases SUV\_bias.
This suggests that excessive frequency domain contrastive regularization may over constrain the reconstruction and weaken spatial domain intensity fidelity.
Although $\lambda_{\mathrm{FCHM}}=0.5$ yields the lowest middle frequency error, $\lambda_{\mathrm{FCHM}}=1.0$ achieves the best overall balance across image fidelity, quantitative consistency, and frequency domain recovery.
Therefore, we set $\lambda_{\mathrm{FCHM}}=1.0$ in all experiments.

\subsection{Comparison with Representative Denoising Methods}

We compare AnF-DiffPET with representative CNN-based, GAN-based, transformer-based, and diffusion-based image recovery methods adapted to PET denoising.
The compared methods include ESRGAN~\cite{esrgan}, VDSR~\cite{vdsr}, EDSR~\cite{edsr}, CDM-GAN~\cite{cdmgan}, SwinIR~\cite{swinir}, I2SB~\cite{I2SB}, and RDDM~\cite{rddm}.
For methods that do not natively support CT conditioning, we provide the registered CT image by stacking it with the LD PET image along the channel dimension. The prediction target remains the raw image.

Table~\ref{compar} summarizes the quantitative comparison across the four PET/CT datasets.
AnF-DiffPET achieves the best overall performance on all datasets, consistently improving PSNR and SSIM while reducing SUV\_bias compared with the strongest competing methods.
For example, on HECKTOR, AnF-DiffPET improves PSNR from 33.30 dB to 36.46 dB and reduces SUV\_bias from 8.28\% to 7.74\% compared with RDDM. Moreover, AnF-DiffPET yields consistently small standard deviations across the evaluated metrics, indicating more stable denoising performance and more consistent quantitative recovery across test samples.
The performance gap heatmap in Fig.~\ref{fig:quantitative_metric_trends} further shows that these gains are consistent across image fidelity and quantification metrics.

Qualitative examples in Fig.~\ref{fig:sota} and Fig.~\ref{fig:external_cases} further support the quantitative findings.
Specifically, Fig.~\ref{fig:sota} presents representative visual comparisons on the HECKTOR dataset, including full-slice PET denoising results and corresponding ROI zoom-in views, which highlight fine-grained anatomical structures and local uptake details.
In contrast, Fig.~\ref{fig:external_cases} shows qualitative results on the remaining datasets (PSMA, NaF, and RIDER), together with residual error maps, demonstrating the robustness of AnF-DiffPET across different anatomical regions and tracer distributions.
Across all datasets, AnF-DiffPET produces clearer anatomical boundaries, more faithful local uptake patterns, and reduced residual artifacts compared with competing methods. These results indicate that the proposed method consistently improves PET image quality while preserving anatomy-consistent uptake structures across diverse PET/CT imaging scenarios.

\subsection{Ablation Study}

\subsubsection{Sequential component ablation}

\begin{table}[!t]
\centering
\caption{Component ablation of AnF-DiffPET on HECKTOR. Values are reported as mean $\pm$ standard deviation, and SSIM and SUV\_bias are reported in percentage.}
\label{tab:ablation_short}
\scriptsize
\setlength{\tabcolsep}{4pt}
\renewcommand{\arraystretch}{1.15}
\resizebox{\linewidth}{!}{%
\begin{tabular}{llcccccc}
\toprule
\multirow{2}{*}{Method} & \multirow{2}{*}{Variant}
& \multicolumn{3}{c}{Components}
& \multirow{2}{*}{PSNR~(dB)~$\uparrow$}
& \multirow{2}{*}{SSIM~(\%)~$\uparrow$}
& \multirow{2}{*}{SUV\_bias~(\%)~$\downarrow$} \\
\cmidrule(lr){3-5}
& & AFG & MSCTR & FCHM & & & \\
\midrule
Baseline
& -- & & &
& $33.30\pm0.24$ & $91.60\pm0.15$ & $8.28\pm0.14$ \\
+ AFG
& M1 & \checkmark & &
& $35.16\pm0.27$ & $94.10\pm0.10$ & $8.11\pm0.11$ \\
+ AFG + MSCTR
& M2 & \checkmark & \checkmark &
& $35.93\pm0.24$ & $94.60\pm0.14$ & $7.86\pm0.09$ \\
+ AFG + FCHM
& M3 & \checkmark & & \checkmark
& $35.54\pm0.21$ & $94.50\pm0.15$ & $8.01\pm0.09$ \\
\midrule
\textbf{Ours}
& M4 & \checkmark & \checkmark & \checkmark
& $\mathbf{36.46\pm0.23}$ & $\mathbf{95.10\pm0.11}$ & $\mathbf{7.74\pm0.10}$ \\
\bottomrule
\end{tabular}
}
\end{table}

\begin{figure}[htbp]
    \centering
    \includegraphics[width=\linewidth]{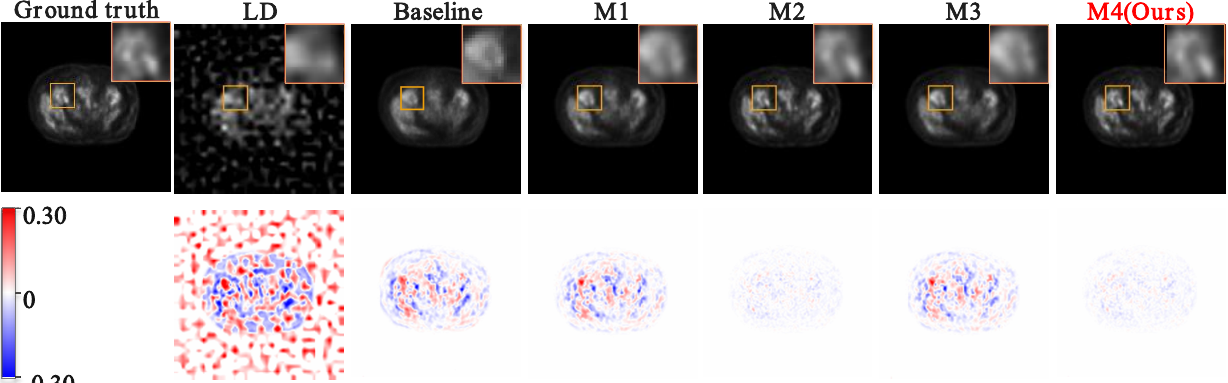}
    \caption{Representative ablation case with PET visualization. Columns show the ground truth HD PET image, LD PET input, the baseline model, and variants M1--M4, where M1--M4 correspond to +AFG, +AFG+MSCTR, +AFG+FCHM, and the complete AnF-DiffPET model, respectively. The upper row presents the denoised PET image with ROI zoom-in views (top-right insets) for detailed uptake visualization, whereas the lower row shows the signed residual error map computed as prediction minus ground truth. Red regions indicate overestimated uptake, whereas blue regions indicate underestimated uptake.}
    \label{fig:ab}
\end{figure}

Table~\ref{tab:ablation_short} summarizes the component wise ablation results.
Starting from the vanilla diffusion U-Net baseline method, adding AFG improves reconstruction fidelity by introducing CT derived anatomical guidance, while MSCTR further improves performance by stabilizing multi-scale decoder feature reconstruction.
Adding FCHM provides additional gains by improving frequency domain consistency during training.
The full method achieves the best overall PSNR and SSIM and the lowest SUV\_bias, indicating that AFG, MSCTR, and FCHM provide complementary contributions to anatomy aware and quantitatively consistent LD PET denoising. The standard deviations of the full method are also among the smallest across metrics, suggesting more stable reconstruction performance across test samples.

Fig.~\ref{fig:ab} provides visual evidence for these quantitative gains.
The baseline method produces blurred reconstructions with loss of anatomical detail, whereas AFG and MSCTR progressively restore structural integrity.
Adding FCHM further improves frequency domain consistency, yielding reconstructions that are more consistent with the ground truth, with clearer uptake boundaries and fewer artifacts.

\begin{figure}[htbp]
    \centering
    \setlength{\abovecaptionskip}{2pt}
    \setlength{\belowcaptionskip}{0pt}
    \includegraphics[width=0.8\linewidth]{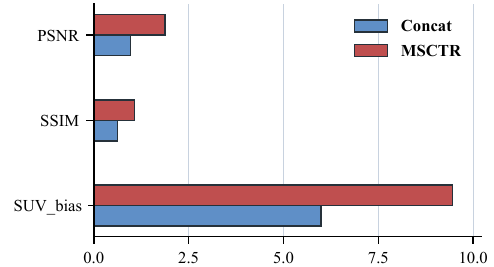}
    \caption{Grouped bar visualization for isolating MSCTR on HECKTOR. Direct feature passing is used as the reference. Bars report relative improvements over Direct, where PSNR and SSIM are measured by relative increases and SUV\_bias is measured by relative reduction.}
    \label{fig:msctr_skip_fusion}
\end{figure}

Because Table~\ref{tab:ablation_short} includes MSCTR within the M2 variant, Fig.~\ref{fig:msctr_skip_fusion} further isolates the skip fusion design.
In this comparison, AFG is enabled and FCHM is disabled, with direct feature passing used as the zero gain reference.
Concat replaces MSCTR with channel concatenation under the same setting, while MSCTR corresponds to the M2 variant in Table~\ref{tab:ablation_short}.
MSCTR increases PSNR to 35.93 dB and SSIM to 94.60\%, while reducing SUV\_bias to 7.86\%.
Compared with Concat, the larger gains of MSCTR show that the proposed module improves feature reuse beyond simple channel concatenation, reducing decoder stage feature drift and improving quantitative consistency.

\FloatBarrier

\section{Conclusion and Discussion}

This paper presents AnF-DiffPET, an anatomy- and frequency-guided diffusion framework for CT conditioned LD PET denoising.
The method integrates AFG, MSCTR, and FCHM to stabilize diffusion-based LD-to-HD PET denoising in both structural and frequency domains.
Across four PET/CT datasets, AnF-DiffPET consistently improves image fidelity, anatomical consistency, and quantitative fidelity over representative CNN-based, GAN-based, transformer-based, and diffusion-based methods.
Ablation and sensitivity analyses further show that AFG, MSCTR, and FCHM contribute complementary benefits.
These findings indicate that coupling anatomical priors with frequency aware diffusion provides an effective strategy for denoising count limited PET.

Although the current evaluation demonstrates stable performance across multiple PET/CT cohorts, further validation under more diverse imaging conditions would be valuable. 
In particular, future studies can include additional scanner models, reconstruction protocols, tracer types, and clinically acquired LD/HD PET pairs to further assess the robustness of the proposed framework. Future work can also extend the current two-dimensional formulation toward 2.5D or three-dimensional denoising, enabling more explicit modeling of through-plane anatomical continuity and volumetric uptake patterns. 
In addition, lesion-level quantitative analysis, organ-level uptake evaluation, and reader-oriented assessment may further complement image-level metrics and better characterize the clinical relevance of denoised PET details.

\bibliographystyle{elsarticle-num}
\bibliography{ref}

\end{document}